\pdfoutput=1

\documentclass[11pt]{article}

\usepackage[]{naacl2021}

\usepackage{times}
\usepackage{latexsym}

\usepackage[T1]{fontenc}

\usepackage[utf8]{inputenc}

\usepackage{microtype}
\usepackage{adjustbox}
\usepackage{booktabs}
\usepackage{multirow}
\usepackage{amsmath}
\usepackage{amssymb}
\usepackage{ulem}

\def\figref#1{Fig.~\ref{#1}}
\def\secref#1{Sec.~\ref{#1}}
\def\tabref#1{Table~\ref{#1}}
\def\eqnref#1{Eqn.~\ref{#1}}

\title{Enriching Transformers with Structured Tensor-Product Representations for Abstractive Summarization}

\author{
 Yichen Jiang\thanks{\:\:Work partially done while at Microsoft Research.}$^{\ \ 1}$,
 Asli Celikyilmaz$^{2}$,
 Paul Smolensky$^{2,3}$,
 Paul Soulos\footnotemark[1]$^{\ \ 3}$,
 Sudha Rao$^{2}$, \\
 \textbf{
 Hamid Palangi$^{2}$,
 Roland Fernandez$^{2}$,
 Caitlin Smith\footnotemark[1]$^{\ \ 3}$,
 Mohit Bansal$^{1}$,
 Jianfeng Gao}$^{2}$ \vspace{5pt} \\
 $^{1}$UNC Chapel Hill~~~~~$^{2}$Microsoft Research, Redmond~~~~~$^{3}$Johns Hopkins University \\
 \small\texttt{ \{yichenj, mbansal\}@cs.unc.edu} \\
 \small\texttt{\{aslicel,psmo,sudha.rao,hpalangi,rfernand,jfgao\}@microsoft.com}\\
 \small\texttt{ \{psoulos1, csmit372\}@jhu.edu} \\
}

\begin{document}
\maketitle

\begin{abstract}
Abstractive summarization, the task of generating a concise summary of input documents, requires: (1) reasoning over the source document to determine the salient pieces of information scattered across the long document, and (2) composing a cohesive text by reconstructing these salient facts into a shorter summary that faithfully reflects the complex relations connecting these facts.
In this paper, we adapt \textsc{TP-Transformer}~\cite{schlag2019enhancing}, an architecture that enriches the original Transformer~\citep{vaswani2017attention} with the explicitly compositional Tensor Product Representation (TPR), for the task of abstractive summarization.
The key feature of our model is a structural bias that we introduce by encoding two separate representations for each token to represent the syntactic structure (with \textit{role vectors}) and semantic content (with \textit{filler vectors}) separately. 
The model then binds the role and filler vectors into the TPR as the layer output.
We argue that the structured intermediate representations enable the model to take better control of the contents (salient facts) and structures (the syntax that connects the facts) when generating the summary.
Empirically, we show that our \textsc{TP-Transformer} outperforms the Transformer and the original \textsc{TP-Transformer} significantly on several abstractive summarization datasets based on both automatic and human evaluations. 
On several syntactic and semantic probing tasks, we demonstrate the emergent structural information in the role vectors and
improved syntactic interpretability in the TPR layer outputs.\footnote{Code and models are available at \\ \url{https://github.com/jiangycTarheel/TPT-Summ}}

\end{abstract}
\section{Introduction}
\begin{figure}[t!]
\begin{center} 
\includegraphics[width=0.48\textwidth]{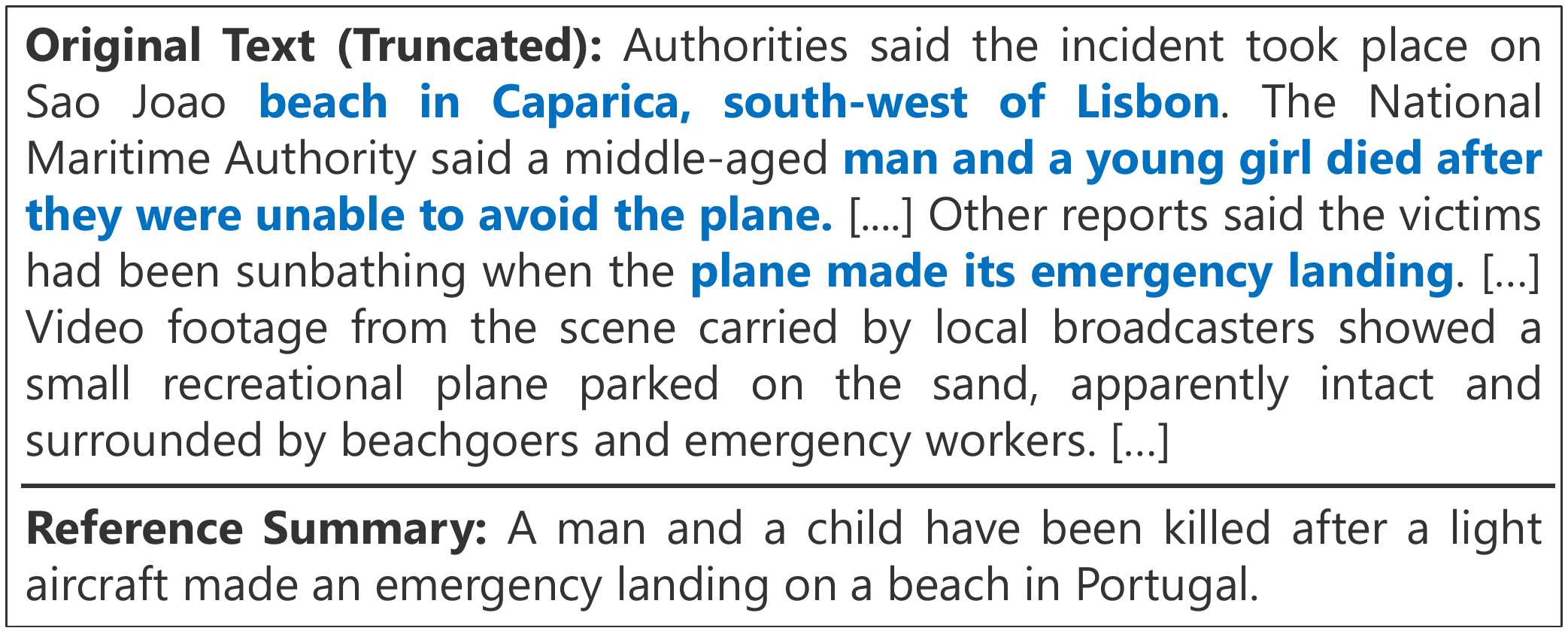}
\end{center} 
\vskip -0.15in
\caption{\small An  example document and its one line summary from XSum dataset. Document content that is composed into an abstractive summary is color-coded.\vspace{-15pt}}
\label{fig:firstpage}
\end{figure}

Abstractive summarization is the task of generating a shorter version of a source text
without necessarily reusing the sentences from the original source, while preserving the 
meaning of its salient contents. 
It is a complex task that requires: semantic understanding of the source text and reasoning over its lexical units, making inferences about their relation to extract salient facts which are scattered across the long document, as well as generating a concise and coherent sequence of new sentences that covers the salient facts. 
While humans are remarkably good at this type of reasoning and abstraction, 
developing models that are capable of extraction, comprehension, abstraction, and reformulation of salient contents has been an open research question.

One prominent aspect of abstractive summarization is that models struggle with combining multiple salient aspects in the source text into a coherent and grammatical set of sentences that preserve the original information in the source document.
As shown in \figref{fig:firstpage}, these pieces of salient information (``death", ``emergency landing", ``beach") are often connected by complex syntactic, causal, and temporal relations and are loosely grouped under the main topic of the source document.
The transformer models~\cite{vaswani2017attention} 
encode syntactic and semantic information of the input text into a single representation space with the self-attention, 
and decode the salient aspects into a short summary with the cross-attention. 
However, despite the large number of training examples, 
current state-of-the-art transformer based approaches still struggle with systematic generalization of the composition of multiple salient pieces of information.

\begin{figure}[t!]
\begin{center} 
\includegraphics[width=0.45\textwidth]{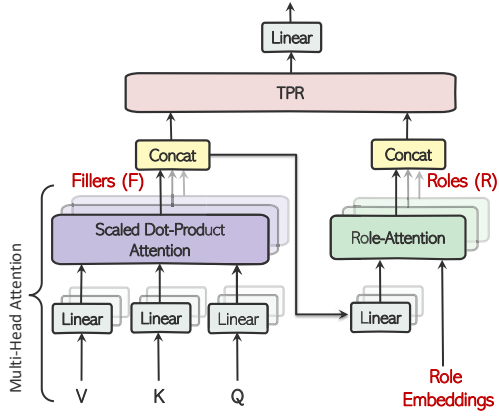}
\end{center} 
\caption{\small The Filler and Role Binding operation of the \textsc{\textsc{TP-Transformer}} Model architecture.\vspace{-10pt}}
\label{fig:TPR}
\end{figure}

In this paper, we investigate new types of computational primitives for transformers based on Tensor Product Representations (TPRs)~\cite{smolensky1990tpr} which are explicitly-compositional vector embeddings of symbolic structures.
A Tensor Product Representation encodes a constituent in a symbolic structure as a composite 
of a \textit{role}, which encodes the structural information (e.g., the dependency relation with another word), and a \textit{filler}, which encodes the content of the constituent (e.g., the meaning of a word). 
Analogously, the \textsc{TP-Transformer} constructs a pair of representations for every token at every layer: a \textit{filler vector} returned by attention 
and a novel \textit{role vector}.
As visualized in \figref{fig:TPR}, the model then binds the role and filler vectors
to produce the output of every token as a TPR.
We adapt the \textsc{TP-Transformer}~\cite{schlag2019enhancing}, which was proposed for solving mathematics problems, for the task of abstractive summarization.
Unlike the original \textsc{TP-Transformer}, which directly projects the input representation into a \textbf{continuous} role vector space, 
our model generates the role vectors by attending to a learned dictionary of role embeddings~\cite{Palangi2018QuestionAnsweringWG}.
We observe that most learned role attention distributions are approximately one-hot, thus restricting the role vectors to a highly \textbf{discrete} space. 
This structural inductive bias encourages the \textsc{TP-Transformer} to encode the syntactic information in the discrete roles while isolating the semantics in the continuous fillers. 

To test the ability of our \textsc{TP-Transformer} with discrete roles against the standard Transformer and the \textsc{TP-Transformer} with continuous roles,
we build several models from scratch on a number of summarization datasets spanning different degrees of abstractiveness, output summary lengths, and domains.
Our \textsc{TP-Transformer} significantly outperforms the standard Transformer and the \textsc{TP-Transformer} with continuous roles on the XSum~\cite{narayan-etal-2018-dont}, Wikihow~\cite{koupaee2018wikihow}, and Arxiv~\cite{cohan-etal-2018-discourse} datasets and achieves competitive performance on the CNN/Daily Mail~\cite{hermann2015teaching,nallapati2016abstractive} dataset, measured by automatic metrics including ROUGE~\cite{Lin-04-rouge} and METEOR~\cite{Denkowski-14-meteor}.
Our human evaluations on XSum and Wikihow datasets also correlate with the automatic metrics, demonstrating that summaries generated by our \textsc{TP-Transformer} are indeed better than the Transformer's generations.

Furthermore, to investigate the structural representation that naturally emerges during training and the advantage of having compositional TPR hidden states, 
we design a suite of decoder probing tasks to explore the information encoded in the role, filler, and TPR space.
We adopt the encoder probing task design presented in~\citet{tenney2018edge} and create four decoder probing tasks: Part-of-speech tagging (POS), Dependency Labeling (DEP), Semantic Role Labeling (SRL), and Named Entity Labeling (NEL).
Our findings collectively show that the decoder's role vectors encode a wealth of syntactic structures, aiding the decoder in deducing the syntactic features (e.g., being a proper noun, being the object of the root predicate) of the next token to be generated.
The decoder's filler vectors on the other hand encode more semantic information (e.g., being a person's name). 
Furthermore, we observe that having the compositional TPR results in a more interpretable final representation than the original Transformer has at every layer, regarding the syntactic features of the next word to be generated.
Our results support our hypothesis that by disentangling semantics and syntax, such structured intermediate representations enable the model to better control both the content to be conveyed and the syntactic structure needed to express it, ultimately improving the factuality and grammaticality of the generated summaries. 

Our overall contributions are as follows: (1) we present a novel adaptation of the original Transformer architecture that incorporates a dictionary of role embeddings at every layer and generates Tensor Product Representation by binding the role vectors with attention outputs (filler vectors); (2) show that our \textsc{TP-Transformer} outperforms the Transformer as well as the original \textsc{TP-Transformer}~\cite{schlag2019enhancing} on several abstractive summarization datasets; and (3) demonstrate the emergent structures in representations by revealing the disentangled syntactic and semantic information encoded in the role and filler spaces.
\section{The \textsc{TP-Transformer}}
\label{sec:model}

We build our \textsc{TP-Transformer} based on the Transformer architecture used in~\citet{raffel2020T5}.
A \textsc{TP-Transformer} encoder applied to a sequence of tokens $i = 1, ..., I$ can be seen as a 2-dimensional lattice of cells $(i, l)$ where $i$ is the position of the input token and $l = 1, ..., L$ are the layer indices. 
All cells in the encoder have the same architecture and the cells at the same layer share the same weights. 
We introduce the basic components of a \textsc{TP-Transformer} cell in \secref{ssec:TPT_cell} and its encoder and decoder cells in \secref{ssec:TPT_enc_dec}.

\subsection{Tensor-Product Representation Basics}
Tensor-Product Representations (TPR;~\cite{smolensky1990tpr}) are explicitly-compositional vector embeddings of symbolic structures, where each constituent of the structure is represented as the product of a \textbf{role} vector, which encodes its structural information, and a \textbf{filler} vector, which contains the content.
The TPR of a whole structure is the sum of the representation of its constituents.
To represent any 3-digit number using TPRs, we need three role vectors: $\{r(p1)$: Ones place, $r(p2)$: Tens place, $r(p3)$: Hundreds place$\}$ and ten filler vectors $f$ for ten digits. 
For example, the TPR of the number 985 is $r(p1)\otimes f(5) + r(p2)\otimes f(8) + r(p3)\otimes f(9)$, where $\otimes$ is the tensor product.
When representing a number, the role vectors operate similarly as the positional embeddings in a Transformer~\cite{vaswani2017attention}.
However, when representing natural languages, the role vectors need to encode a variety of structural information (e.g., predicate-argument, tense, etc) and thus it is infeasible to hand-design an entire suite of role vectors as we did for numbers.
To overcome this challenge, for every token, we dynamically compute its role \textit{vector} from a dictionary of a finite number of role \textit{embeddings} learned with the entire model and treat the self-attention outputs as the fillers.
We introduce the full computation procedure in \secref{sssec:TPR}.

\subsection{The \textsc{TP-Transformer} Cell}
\label{ssec:TPT_cell}
Similar to the basic Transformer cell, at every layer, a \textsc{TP-Transformer} Encoder cell starts with a layer normalization and the multi-head self-attention followed by a residual layer.
Then, the cell treats the output vectors as fillers and binds them to role vectors to construct a Tensor Product Representation, which is then passed through the feed-forward network to yield the final states.

\subsubsection{Multi-Head Attention}
The \textsc{TP-Transformer} cell adopts multi-head attention~\cite{vaswani2017attention} to enable information passing between tokens.
At any layer, denote the input vectors as $X$$\in$$\mathbb{R}^{k_x\times d_m}$ and the attention target vectors as $Y$$\in$$\mathbb{R}^{k_y\times d_m}$, where $k_x, k_y$ are the length of the sequences and $d_m$ is the dimension of the input vectors.
In the case of self attention, we have $Y$=$X$; while for the encoder-decoder cross attention, $Y$ is the encoder's output vectors.
We first apply layer normalization~\cite{ba2016layer} to get $\hat{X}$ and then linearly project it to the query, key, and value vectors for each attention head $h = 1, ..., H$.
\begin{equation} \label{eq:attention_qkv}
\begin{split}
Q^h &= \hat{X}\mathbf{W}_q^h  + \mathbf{b}_q^h\\
K^h &= Y\mathbf{W}_k^h + \mathbf{b}_k^h\\
V^h &= Y\mathbf{W}_v^h + \mathbf{b}_v^h\\
\end{split}
\end{equation}
where $\mathbf{W}_q, \mathbf{W}_k, \mathbf{W}_v \in \mathbb{R}^{d_m \times d_k}$. The attention output matrix $\Bar{V}$ for each head $h$ is computed as: 
\begin{equation} \label{eq:attention}
\bar{V} = \mathrm{softmax}(\frac{QK^T}{\sqrt{d_k}})V
\end{equation}
where $d_k$ is the dimension of the key vectors $K$. 
The multi-head attention output $O$ is the concatenation of the attention outputs from all heads followed by another linear projection $\mathbf{W}_o \in \mathbb{R}^{d_m \times d_m}$.
We end the Multi-head Attention with a residual connection with the layer input vectors $\hat{X}$:
\begin{equation} \label{eq:attention_concat}
\begin{split}
\mathrm{MHAttn}(X,Y) &= \hat{X} + [\Bar{V_1}, ..., \Bar{V_H}] \mathbf{W}_o
\end{split}
\end{equation}
where $\bar{V}_h$ is the attention output for the $h$-th head.

\subsubsection{Computing TPRs}
\label{sssec:TPR}
\paragraph{Role Embeddings.}
Following \citeauthor{Palangi2018QuestionAnsweringWG} (\citeyear{Palangi2018QuestionAnsweringWG}),
but departing from \citeauthor{schlag2019enhancing} (\citeyear{schlag2019enhancing}), every layer of our \textsc{TP-Transformer} is equipped with a dictionary $\mathbf{r} \in \mathbb{R}^{N_r \times d_r}$ of $N_r$ distinct role embeddings with a dimension of $d_r$. 
Each role embedding $\mathbf{r}_n$, $n$=1,$\dots$,$N_r$, is randomly initialized in the entire network. 
The role embeddings are normalized before computing role vectors:
\begin{equation}
\label{eq:norm_role}
\hat{\mathbf{r}}_n = \frac{\mathbf{r}_n}{\lVert \mathbf{r}_n \rVert_2}\; \mathrm{for}\; n = 1, ..., N_r \\
\end{equation}
At each layer, the model computes a weighted combination of these role embeddings $\hat{\mathbf{r}}$ to form a unique \textit{role vector} for every token.

\paragraph{Multi-Head TPR Binding.}
Our \textit{filler vectors} correspond to the multi-head attention output $F = \mathrm{MHAttn}(X)$ (\eqnref{eq:attention_concat}).
The filler $F$ of each token has a corresponding role vector $R$.
We first compute the $R^h \in \mathbb{R}^{d_r}$ at every head $h = 1, ..., H$ as a weighted average of the normalized role embeddings $\hat{\mathbf{r}}$. 
We then concatenate the $R^h \in \mathbb{R}^{k_x\times d_r}$ of $H$ heads to get the multi-head role vectors $R \in \mathbb{R}^{k_x\times (d_r\cdot H)}$ for all $k_x$ tokens. We define this process formally as:
\begin{equation} \label{eq:role_attn}
\begin{split}
R^h &= \mathrm{softmax}(F\mathbf{W}^h_r) \hat{\mathbf{r}}\\
R &= [R^1, ..., R^H]
\end{split}
\end{equation}
where $\mathbf{W}_r \in \mathbb{R}^{d_m\times N_r}$ is the linear projection that computes the attention scores over the role embeddings for every token.\footnote{We set $d_r \cdot H = d_m$ so that the multi-head role vectors $R$ have the same dimension as $F$.}

We use a Hadamard product\footnote{The Hadamard (or elementwise) product is the diagonal of the full tensor product.} to approximate the full Tensor product in binding the role vectors $R$ with filler vectors $F$, as it was shown in \citet{schlag2019enhancing} that using the Hadamard products allows learning an optimial lower-rank approximation of the full TPRs.
The binding operation is followed by an addition with the unbound fillers ($F$) to return the residual TPR vectors.
\begin{equation} \label{eq:binding}
\mathrm{TPR}(F) = R \odot F + F
\end{equation}

\subsubsection{Residual Feed-forward Layer}
The feed-forward layer of a cell consists of a linear projection followed by a ReLU activation and a second linear projection.
The feed-forward output is then added to the input vectors:
\begin{equation} \label{eq:ff}
\mathrm{FF}(X) = X + \mathrm{ReLU}(X\mathbf{W}_{g} + \mathbf{b}_{g})\mathbf{W}_{f} + \mathbf{b}_{f}\\
\end{equation}
Here, $\mathbf{W}_{g}$$\in$$\mathbb{R}^{d_m \times d_f}$, $\mathbf{b}_{g}$$\in$ $\mathbb{R}^{d_f}$, $\mathbf{W}_{f}$$\in$ $\mathbb{R}^{d_f \times d_m}$, $\mathbf{b}_{f} \in \mathbb{R}^{d_m}$, and $x$ is the function argument.

\subsection{\textsc{TP-Transformer} Encoder \& Decoder}
\label{ssec:TPT_enc_dec}
Given the components of our basic \textsc{TP-Transformer} cell in the previous section, we now describe how we construct the \textsc{TP-Transformer} encoder and decoder.

First, the self-attention and the encoder-decoder cross-attention for every token can be computed as:
\begin{equation} \label{eq:TP-self-cross}
\begin{split}
\mathrm{Self}(X) &= \mathrm{TPR}(\mathrm{MHAttn}(X,X)) \\
\mathrm{Cross}(Y,H) &= \mathrm{TPR}(\mathrm{MHAttn}(Y,H))
\end{split}
\end{equation}
where $H$ is the output of the encoder's final layer.
$Y$ represent the previous layer's output vectors of either the partially (so-far) decoded sequence at test time or the masked reference summary at training time.
The encoder and decoder's operations at every layer can be summarized as:
\begin{equation} \label{eq:TP-enc-dec}
\begin{split}
\mathrm{Encode}(X) &= \mathrm{FF}(\mathrm{Self}(X)) \\
\mathrm{Decode}(H,Y) &= \mathrm{FF}(\mathrm{Cross}(\mathrm{Self}(Y), H))
\end{split}
\end{equation}
After $L$ layers of encoding and decoding, 
the final distribution of the $i$-th output token is given by:
\begin{equation} \label{eq:output_prob}
\hat{z}_i = \mathrm{softmax}(\mathbf{E}^T y_{i,L})
\end{equation}
where $Y_L = \mathrm{Decode}(H,Y_{L-1})$ are the decoder's output states at the last layer and $\mathbf{E}$ is the tied input/output word embeddings.

\section{Summarization Experiments}
\label{sec:exp}

\begin{table*}[t]
\centering
\begin{scriptsize}
\begin{tabular}{p{1.2cm}|p{13.2cm}}
\toprule
 \centering Datasets & Summary \\
\midrule
 \centering XSum & Luxury fashion designer Burberry has returned to profit after opening \underline{new stores} and spending more on online marketing. \\
\midrule
\centering \multirow{2}{*}{\centering Wikihow} & Build a trustworthy bond with \underline{your piggy}. Research different \underline{training methods}. Choose the \underline{training method} that works best for you and \underline{your guinea pig}. Gather the materials that \underline{you will need} for training. \\
\midrule
 \centering Arxiv\newline (Abbreviated) & We study the \underline{phase behavior of a nematic liquid crystal} \underline{confined between} a flat substrate with strong anchoring and a \underline{patterned substrate} whose structure and local anchoring strength we vary. 
[\dots]
In addition the effective energy method allows one to determine the \underline{energy barriers between} two states in a \underline{bistable nematic device} .\\
\midrule
 \centering \multirow{2}{*}{\centering CNN/DM} & \underline{Mentally ill inmates} in Miami are \underline{housed on the "forgotten floor"}. \underline{Judge Steven Leifman says} most are there as a result of \underline{"avoidable felonies"}. While CNN tours facility, patient shouts: \underline{"I am the son of the president"}. \\
\bottomrule
\end{tabular}
\vspace{-5pt}
\caption{\small Example summaries from XSum, Arxiv, Wikihow, and CNN/Daily Mail datasets. Text segments directly extracted from the source document are underlined.
}
\label{table:datasets}
\end{scriptsize}
\end{table*}

\begin{table*}[t]
\centering
\begin{small}
\begin{tabular}[t]{ccc|cc|c}
\toprule
Datasets & Split &\# beam & Transformer & TPT-c~\cite{schlag2019enhancing} & TPT-d (Ours) \\
\midrule
\multirow{4}{*}{XSum} & \multirow{2}{*}{Dev}
 & 1 &  33.34/12.07/26.47/22.28 & 30.73/10.38/24.39/21.14 & \textbf{34.61}/\textbf{13.13}/\textbf{27.59}/\textbf{23.43} \\
& & 4 & 34.48/13.08/27.29/24.59 & 31.83/11.28/25.11/22.39 & \textbf{35.70}/\textbf{14.11}/\textbf{28.38}/\textbf{25.80}\\
& \multirow{2}{*}{Test} 
  & 1 & 33.22/11.90/26.32/23.02 & 30.74/10.23/24.32/21.11 & \textbf{34.62}/\textbf{12.98}/\textbf{27.49}/\textbf{24.38}\\
& & 4 & 34.46/12.97/27.21/24.42 & 32.01/11.26/25.19/22.45 & \textbf{35.84}/\textbf{14.06}/\textbf{28.40}/\textbf{25.79}\\
\midrule
\multirow{4}{*}{Wikihow} & \multirow{2}{*}{Dev}
  & 1 & 33.11/11.90/25.46/19.00 & 28.44/7.65/20.07/16.38 & \textbf{34.12}/\textbf{12.36}/\textbf{26.02}/\textbf{20.16} \\
& & 4 & 35.85/13.32/26.83/21.57 & 29.98/8.34/20.70/17.95 & \textbf{36.54}/\textbf{13.69}/\textbf{27.21}/\textbf{22.53}\\
& \multirow{2}{*}{Test} 
  & 1 & 33.40/12.18/25.66/19.31 & 28.63/7.82/20.23/16.49 & \textbf{34.19}/12.47/\textbf{25.99}/\textbf{20.23} \\
& & 4 & 35.91/13.49/27.01/21.57 & 30.13/8.50/20.78/18.11 &  \textbf{36.70}/13.75/\textbf{27.36}/\textbf{22.53}\\
\midrule
\multirow{4}{*}{Arxiv} & \multirow{2}{*}{Dev}
  & 1 & 35.08/10.13/31.86/19.91 & 32.27/7.50/29.34/17.72 & \textbf{35.91}/\textbf{10.32}/\textbf{32.55}/\textbf{20.82} \\
& & 4 & 37.95/11.48/34.03/23.31 & 34.45/8.40/30.91/20.17 & \textbf{38.35}/11.56/\textbf{34.32}/\textbf{23.74}\\
& \multirow{2}{*}{Test} 
  & 1 & 35.00/9.98/31.79/19.72  & 32.46/7.53/29.47/17.75 & \textbf{35.82}/\textbf{10.12}/\textbf{32.46}/\textbf{20.65}\\
& & 4 & 38.01/11.33/34.02/23.19 & 34.68/8.50/31.15/20.17 & \textbf{38.36}/11.43/\textbf{34.29}/\textbf{23.61}\\
\midrule
\multirow{4}{*}{CNN/DM} & \multirow{2}{*}{Dev}
  & 1 & 40.56/18.18/37.73/31.91 & 39.66/17.45/36.99/31.15 & 40.61/18.17/31.77/31.35 \\
& & 4 & 41.97/19.23/38.84/34.55 & 41.49/18.83/38.45/34.14 & 41.81/19.11/38.73/34.49 \\
& \multirow{2}{*}{Test} 
  & 1 & 39.83/\textbf{17.63}/37.02/31.75 & 39.10/16.96/36.41/31.15 & 39.63/17.35/36.80/31.57 \\
& & 4 & 41.22/\textbf{18.70}/38.09/34.50 & 40.68/18.19/37.70/33.99 & 41.01/18.38/37.91/34.34 \\
\bottomrule
\end{tabular}
\vspace{-5pt}
\caption{\small Automatic evaluation results on the dev/test set of \textbf{XSum}, \textbf{Arxiv}, \textbf{Wikihow}, and \textbf{CNN/Daily Mail} dataset. The results in every cell represent \textbf{F1} variant of \textbf{ROUGE-1}/\textbf{ROUGE-2}/\textbf{ROUGE-L}/\textbf{METEOR} scores. The best ROUGE scores with a statistically significant advantage, and the best METEOR scores with at least 0.3 advantage are bolded.
 \vspace{-10pt}
}
\label{table:summ_results}
\end{small}
\end{table*}

\begin{table*}[t]
\centering
\begin{small}
\begin{tabular}[t]{cc|ccccc|c}
\toprule
Datasets & Models & Grammar & Coherency & Faithfulness & Saliency & Repetition & Overall \\
\midrule
\multirow{3}{*}{XSum} & 
Transformer wins             & 39 & 48 & 43 & \textbf{50} & 38 & 48\\
&\textsc{TP-Transformer} wins & \underline{\textbf{47}} & 48 & \textbf{46} & 47 & \textbf{42} & \textbf{52}\\
&Tie / No agreement            & 34 & 24 & 31 & 23 & 40 & 20\\
\midrule
\multirow{3}{*}{Wikihow} & 
Transformer wins              & 45 & 45 & 43 & \underline{\textbf{54}} & 48 & 43 \\
&\textsc{TP-Transformer} wins & \textbf{48} & 45 & \textbf{46} & 47 & 48 & \underline{\textbf{59}} \\
&Tie / No agreement             & 27 & 30 & 31 & 19 & 24 & 18 \\
\bottomrule
\end{tabular}
\vspace{-5pt}
\caption{\small Human Evaluation results on 120 random samples from the XSum~\cite{narayan-etal-2018-dont} and Wikihow~\cite{koupaee2018wikihow} test sets. 
The best numbers with an advantage of at least 5 points are underlined. 
 \vspace{-10pt}
}
\label{table:human_eval}
\end{small}
\end{table*}

\subsection{Abstractive Summarization Datasets}
\label{ssec:summ_data}
We train our models on four English abstractive summarization datasets varying the level of abstractiveness (explained below) and the length of summaries, as well as input domain.

\paragraph{XSum} \cite{narayan-etal-2018-dont} consists of 227k BBC articles from 2010 to 2017 concerning various subjects along with professionally written single-sentence summaries.
Its summaries cover a wide variety of syntactic structures (relative clause, etc) and relations (causal, temporal, etc).

\paragraph{Wikihow} \cite{koupaee2018wikihow} is a dataset consisting of instructions from the WikiHow.com website. 
Each of 200k examples has multiple instruction-step paragraphs, each paired with a summarizing sentence. 
The task is to generate the concatenated summaries of all paragraphs.

\paragraph{Arxiv} \cite{cohan-etal-2018-discourse} is a long document summarization dataset of scientific publications from arXiv.org (113k).
The task is to generate the abstract from the paper body.

\paragraph{CNN/Daily Mail} \cite{hermann2015teaching,nallapati2016abstractive} dataset contains 93k articles from CNN and 220k articles from the Daily Mail. 
Every article is accompanied by a few human-written bullet points about its content.
We use the non-anonymized version used in \citet{see-etal-2017-get}.

\paragraph{Dataset Abstractiveness.} 
We show a summary from each of these four datasets in \tabref{table:datasets}.
According to the comparison made by \citet{zhang2020pegasus} using the coverage and density measures~\cite{grusky2018newsroom}, the XSum and Wikihow datasets are more abstractive than the others since their summaries rarely contain large chunks of words overlapping with the source documents.
CNN/Daily Mail is the least abstractive of the four.
Furthermore, in most cases, a sentence in a CNN/Daily Mail summary only refers to a single sentence from the source document as suggested in~\citet{lebanoff-etal-2019-scoring}, while a sentence in an XSum or Wikihow summary usually aggregates information from multiple source sentences. 

\subsection{Experimental Setup}
The Transformer and the two \textsc{\textsc{TP-Transformers}} all have 6 layers, 8 heads per layer, dimension per head $d_k$=64, model dimension $d_m$=512, and feed-forward dimension $d_f$=2048 for the encoder and decoder.
Our \textsc{TP-Transformer} with discrete roles has $N_r$=50 role embeddings of dimension $d_r$=64 at every layer.
For each dataset above, we train the all three models from scratch using an Adafactor Optimizer~\cite{adafactor-shazeer18a} with square root learning rate decay and dropout rate of 0.1.
We evaluate the models using automatic metrics including ROUGE F1 score and METEOR. 

\subsection{Results}
\label{ssec:summ_results}
We report automatic metric scores from our evaluated models in \tabref{table:summ_results}.
We refer to the \textsc{TP-Transformer}, with freely-generated continuous role vectors (no role dictionary)~\cite{schlag2019enhancing} as TPT-c, and our own \textsc{TP-Transformer} with a discrete set of role embeddings as TPT-d.
On the XSum, Arxiv, and Wikihow datasets, our \textsc{TP-Transformer} (TPT-d) outperforms the original Transformer on all metrics.
On the CNN/Daily Mail dataset, both models obtain similar performance across all metrics.
On every dataset, the TPT-c model which excels on the mathematics dataset, is the worst among the three models being compared.
This suggests that continuous role vectors are not suited to the summarization tasks.

As we explain in \secref{ssec:summ_data}, CNN/Daily Mail is the least abstractive one among the four datasets.
In contrast, summaries from the XSum and Wikihow datasets contain very few n-grams (n$>$2) that can be copied from the source documents and thus push the model's ability to compose a coherent summary restating the salient aspects from the source.
Furthermore, as illustrated in \tabref{table:datasets}, 
the XSum summary contains a long sentence that combines multiple pieces of information scattered through the long source document. 
These facts are usually connected by syntactic, temporal\footnote{``returned to profit \textbf{after} opening new stores"}, or causal\footnote{``Opening new stores and spending more on online marketing" caused "more profit".} relations and thus the model must be able to connect and reason across these salient facts and then convert them into a coherent sentence that faithfully reflects the original facts and their relations. 
We argue that the compositional TPR can better enable these abilities required for XSum, where we indeed find that our \textsc{TP-Transformer} achieves the largest advantage over the Transformer among its improvements on all datasets.

\subsection{Human Evaluation}
We conduct human evaluation to compare the summaries generated by the Transformer and our \textsc{TP-Transformer}.
We randomly sample 120 examples from the test sets of XSum and Wikihow datasets with the beam-searched model summaries.
We refer to appendix for the complete setup.
As shown in \tabref{table:human_eval}, on the XSum dataset, summaries generated by the \textsc{TP-Transformer} are significantly better in grammar.
This corroborates our claim that having the TPR can improve the model's ability to follow the correct syntax in composing the summary.
On the Wikihow dataset, the Transformer receives more votes in regarding the saliency.
However, our \textsc{TP-Transformer} maintains an advantage in grammar and achieves significantly better overall preferences.

\paragraph{Unfaithful XSum Examples}
It is well-known that the XSum dataset contains a portion of unfaithful reference summaries that mention facts not included in the source article~\cite{durmus-etal-2020-feqa,maynez-etal-2020-faithfulness}.
Therefore, we are interested to find out whether our \textsc{TP-Transformer} 
is better than the baseline only at expressing the faithful content or it can also generate some external, ``unfaithful" facts that the baseline can't cover.
To answer this question, we randomly sample 100 examples from the XSum dev set and manually examine the source document, reference summary, and the two generated summaries.
Among these 100 examples, we identify 71 examples whose reference summary includes ``unfaithful" facts that are not mentioned in the source. 
In 21 out of 71 examples, the Transformer baseline manages to generate some ``unfaithful" facts that match those in the reference while our \textsc{TP-Transformer} achieves this in 17 examples.
Such ``unfaithful" facts that were recovered by the models include the full name of a person when only the last name is mentioned in the source, the political party or the job title of a person, 
each of which can be attributed to at least one example seen by models during the training.
Therefore, we believe that both models learn to draw external information from its memory of the seen examples,
while our \textsc{TP-Transformer} doesn't do better than the baseline Transformer at referring to external facts to obtain higher ROUGE scores.
\section{Probing Experiments}

Probing is a method to test whether some particular information is present in the model’s encodings.
To achieve this, an auxiliary classifier is trained to predict specified linguistic features from the model’s internal representations.
We probe different components (roles, filler, TPRs) in our \textsc{\textsc{TP-Transformer}}s as well as the attention+residual outputs (equivalent to the filler) of the Transformer to assess the naturally emergent structures encoded in the role vectors and the effectiveness of the TPR in the decoding process.
By conducting the probing experiments, we aim to (1) provide some insights and evidence of the different information encoded by the role and filler vectors; and (2) explain the ROUGE advantage of our \textsc{TP-Transformer} by showing that its output representation can better encode the linguistic structural information concerning multiple probing tasks.

\subsection{Decoder Probing Tasks}
When studying an encoder, previous works probe its $i$-th intermediate representation at a certain layer for information about the $i$-th input token.
For a decoder, however, we probe its $i$-th representation for clues about the $i$-th token it generates given the $i-1$ previously generated tokens as the input.
Intuitively, we are probing for the decoder's internal decision about the syntactic roles and semantic content of this token before it was ultimately selected.
Based on encoder probing tasks used by~\citet{tenney2018edge}, we select and adapt four tasks to probe our decoders.

\paragraph{Part-of-speech tagging (POS)} is the syntactic task of assigning tags such as noun (singular/mass noun: NN, proper noun: NNP, etc), verb (past tense: VBD, past participle: VBN, etc), adjective (comparative: JJR, etc), etc. to each token $i$. 
We let $s_1 = [i, i + 1)$ be a single token, and seek to predict its POS tag.

\paragraph{Dependency labeling (DEP)} seeks to predict the functional relationships of one token relative to another: e.g. is it a modifier-head relationship, a subject-verb relationship, etc. 
We take $s_1 = [i, i + 1)$ to be a single token and $s_2 = [j, j + 1)$ to be its
syntactic head, and seek to predict the dependency relation between tokens $i$ and $j$.

\paragraph{Semantic role labeling (SRL)} is the task of imposing predicate-argument structure onto a sentence.
We let $s_1 = [i_1, j_1)$ represent a known predicate (e.g., ``push") and $s_2 = [i_2, j_2)$ represent a known argument (``Peter") of that predicate, and seek to predict the role that the argument $s_2$ fills–e.g. ARG0 (agent, the pusher) vs. ARG1 (patient, the pushee).

\paragraph{Named entity labeling (NEL)} is the task of predicting the category of an entity. The categories include \textit{PERSON}, \textit{LOCATION}, \textit{ORGANIZATION}, etc. 
We let $s_1 = [i, j)$ represent a known entity span and seek to predict its type.

\begin{table}[t]
\centering
\begin{small}
\begin{tabular}[t]{cc|c|c}
\toprule
Tasks & Layer & Transformer & TPT-d (Ours) \\
\midrule
\multirow{6}{*}{POS} & 
  1 & -/\textbf{58.4}/58.4  &  36.1/57.1/58.2 \\
& 2 & -/\textbf{65.4}/\textbf{65.4}  &  43.6/63.5/64.4 \\
& 3 & -/\textbf{68.6}/68.3  &  50.4/67.4/68.5 \\
& 4 & -/70.7/70.7  &  50.4/70.8/\textbf{72.1} \\
& 5 & -/72.5/72.5  &  53.4/\textbf{73.3}/\textbf{73.9}\\
& 6 & -/73.3/73.3  &  56.0/\textbf{73.9}/\textbf{74.5}\\
\midrule
\multirow{6}{*}{DEP} & 
  1 & -/78.1/78.1  &  53.1/\textbf{78.8}/\textbf{78.9} \\
& 2 & -/85.0/85.0  &  59.9/84.8/84.7\\
& 3 & -/87.1/87.1  &  66.7/87.4/87.3\\
& 4 & -/87.4/87.4  &  62.9/\textbf{88.3}/\textbf{88.2}\\
& 5 & -/85.0/85.0  &  64.8/\textbf{88.3}/\textbf{87.6}\\
& 6 & -/86.1/86.1  &  60.8/\textbf{86.8}/\textbf{86.6}\\
\midrule
\multirow{6}{*}{SRL} & 
  1 & -/78.2/78.2  &  73.1/78.5/78.4 \\
& 2 & -/79.0/79.0  &  73.8/\textbf{79.8}/79.3\\
& 3 & -/79.6/79.6  &  73.8/79.9/80.0\\
& 4 & -/78.7/78.7  &  73.1/\textbf{80.1}/\textbf{80.2}\\
& 5 & -/77.7/77.7  &  72.9/\textbf{79.9}/\textbf{79.8}\\
& 6 & -/78.1/78.1  &  71.8/\textbf{79.2}/78.2\\
\midrule
\multirow{6}{*}{NEL} & 
  1 & -/59.7/59.7  &  33.3/\textbf{61.4}/\textbf{60.8}  \\
& 2 & -/67.6/67.6  &  37.6/\textbf{68.1}/\textbf{68.2} \\
& 3 & -/69.6/69.6  &  41.5/\textbf{70.9}/\textbf{71.0} \\
& 4 & -/71.8/71.8 &   43.6/\textbf{74.3}/\textbf{73.2}\\
& 5 & -/72.3/72.3  &  44.7/\textbf{76.3}/\textbf{75.7}\\
& 6 & -/73.3/73.3  &  42.2/\textbf{76.1}/\textbf{73.8}\\
\bottomrule
\end{tabular}
\vspace{-5pt}
\caption{\small Results (F1 scores) of probing different intermediate representations in decoders trained on XSum dataset. The results in every cell are presented in the order of \textbf{roles}, \textbf{fillers}, and \textbf{final representations}.
The best numbers with an advantage of at least 0.5 F1 scores are bolded.
 \vspace{-10pt}
}
\label{table:probe_results}
\end{small}
\end{table}

\subsection{Experimental Setup}
As there is no existing dataset for probing decoders, we create our own training and evaluation data by running off-the-shelf models on the summarization datasets.
Specifically, to probe a decoder trained on the XSum dataset on the POS task, we run an POS tagger on the reference summaries from the XSum training set and the model-generated summaries for the XSum dev set to create the ground-truth labels for the training set and model-specific dev set. 
We restore the model trained on a summarization dataset and freeze its parameters. 
Following~\citet{tenney2018edge}, we train a span convolution layer followed by a 2-layer MLP on top of the target representation that project it onto the output label space.

\subsection{Results}
\tabref{table:probe_results} presents the results of probing the decoder of a \textsc{TP-Transformer} trained on the XSum~\cite{narayan-etal-2018-dont} dataset.
Note that the Transformer doesn't have role vectors.
It directly outputs the vector after the multi-head attention and the residual layer. 
Therefore, its fillers and final representations are equivalent.

\paragraph{The decoder role vectors can encode grammatical information while the filler vectors represent the semantics.} We first focus on the results of POS tagging probing task.
Overall, we see a trend of increasing scores as the representations get closer to the final step of computing the distribution over the vocabulary.
This implies that, as the computation progresses through the layers, the generated representations are gradually deciding the POS tag of the next word to generate.
Next, we observe that the role vectors (the 1st number in the TPT-d column) of \textsc{TP-Transformer} encode a considerable amount of information about the POS tag of the next word generated.
Additionally, because the job of deducing the POS tag of the next word is partially shared by the role vectors, the filler vectors' performance degrades compared to the Transformer.
This pattern demonstrates that the \textsc{TP-Transformer}'s decoder is representing the next word to be generated as a composite of structural information encoded in the role vectors and semantic contents encoded in the filler vectors.
Comparing the fillers (the 2nd number in TPT-d column) with the TPR (the 3rd number in the TPT-d column) of \textsc{TP-Transformer}, we see that the TPRs, which bind the roles and fillers, outperform the roles and fillers alone at every layer.
This indicates that the TPR effectively aggregates the linguistic knowledge encoded in the roles and fillers into a shared space, where the POS tag of the next word can be decoded more easily than in the role space or filler space alone.
Last, the final representations of \textsc{TP-Transformer} achieve higher F1 scores than their counterparts in the Transformer in the last three layers.
This demonstrates the benefits of having the TPR in interpreting the POS tag of the word to be generated. 

When we consider the Dependency labeling (DEP) and Semantic role labeling (SRL) tasks, we observe that our \textsc{TP-Transformer}'s final representations consistently beat the Transformer across all layers, with only one exception in the DEP task at the layer 2.
We also observe that the \textsc{TP-Transformer}'s advantage becomes larger in the last three layers except for the final layer in SRL task.
However, unlike in the POS task, the TPR only achieve similar F1 scores to the fillers.

Finally, in the Named entity labeling (NEL) task which is considered to require more semantic information rather than syntax, the role vectors' performance is poorer than their performance in the three syntactic tasks.
For example, the \textsc{TP-Transformer}'s final representations at layer 6 obtain similar F1 scores in the POS and NEL tasks (74.5 VS 73.8), but its role vectors only achieve a 42.2 F1 score in the NEL tasks compared to the 56.0 in the POS.
However, even though the role vectors encode little information about the named entity type of the next token to be generated, the TPR still strongly outperforms the Transformer's filler-only representation at every layer.
We argue that although the syntactic information encoded in the role vectors is not enough to predict the correct named entity, it is still a beneficial complement to the knowledge encoded in the distributed filler vectors in certain situations.
For example, whether the subject ``Chanel" refers to a \textit{PERSON} or an \textit{ORGANIZATION} could depend on its syntactic role and its relation to other words in the sentence (e.g., whether it is the subject or object of ``wears'') .

\paragraph{Compositional representations improves interpretability of the representations.} Overall, by probing the different intermediate representations of the \textsc{TP-Transformer} and the Transformer, we show that having the compositional TPR results in more interpretable final representations at every layer regarding the syntactic features of the next word to be generated.
Considering automatic evaluations generated summaries in \secref{ssec:summ_results}, we argue that this compositionality in learned representation and its syntactic interpretability enable the decoder to take better control of the syntactic structure of the generation when assembling multiple distant facts, and thus lead to summaries of better quality.

\subsection{Discrete Role Vectors}
During the training of our \textsc{TP-Transformer} models on the summarization datasets, we observe that most learned role attention distributions are approximately one-hot, as more than 90\% of the role attention distributions (as computed in \eqnref{eq:role_attn}) have a maximum score larger than 0.98.
Because each role vector is the concatenation of $H$ vectors, each selected from $N_r$ role embeddings,
the completely one-hot role attentions will yield $(N_r)^{H}$ possible role vectors.
Therefore, the learned, approximately one-hot role vectors span $(N_r)^{H}$ discrete subspaces, each of which only covers the close proximity of a concatenation of $H$ role embeddings.
This finding indicates that as we represent the role vectors as multi-head attention over a learnable dictionary of role embeddings,
the structural inductive bias: (1) pushes the role vector space to be even more discrete, and (2) induces the syntactic structures encoded in these discrete role vectors.
We also believe there is a connection between the above two effects, as the structural, syntactic information favors a lower-dimensional or even discrete space while the distributed, semantic information favors a higher-dimensional space.
\section{Related Work}

\paragraph{Explicit TPR Structures in Neural Networks} While earlier TPR work based on~\cite{smolensky1990tpr} focused on computability rather than learnability questions, recently TPRs have been incorporated into several recurrent deep learning models in order to solve various NLP tasks including Part-of-Speech tagging, constituency parsing, image captioning~\cite{huang-etal-2018-TPGN,Huang2019ATPL}, question answering~\cite{Palangi2018QuestionAnsweringWG,schlag2018tprrnn}, and natural-to-formal language generation (program synthesis)~\cite{chen2019natural}.
Most recently, TPRs have been introduced into Transformer architectures, starting with \citet{schlag2019enhancing} which introduced the \textsc{TP-Transformer} to improve the performance and interpretability of mathematical problem solving models.
This model generated continuous role vectors by directly projecting from layer inputs, whereas our model indexes from a dictionary of role embeddings to form the role vectors which are shown to reside in a highly discrete space.

\paragraph{Structured Representations for Abstractive Summarization}
Compared to the extractive methods, abstractive summarization 
models usually fail to show extractive properties, and have 
tendency to copy text from the source \citep{see-etal-2017-get,paulus2018, pasunuru2018, celikyilmaz2018}. More recent approaches that use standard transformers deal with this issue by introducing hierarchical structures to encode local and global information separately focusing on only the semantic content \citep{liu2018,liu2019}. To preserve salient source relations and generate abstractive summaries of the source document, previous work infused models with semantic parsers: while \citet{song-etal-2018-structure} introduces a new structure-infused copy mechanism that combines the source syntactic structure with the copy mechanism, \citet{liao-etal-2018-abstract} uses abstract meaning representations (AMR). While these approaches require that the document sentence semantic parsers are provided beforehand, our models can implicitly learn to approximate the syntactic structure and semantic content in their representations.

\section{Conclusion}
In this work, we enrich the Transformer model with the structured Tensor Product Representation for abstractive summarization tasks.
We represent every token as a pair of role and filler vectors.
We show that our \textsc{TP-Transformer} with discrete roles outperforms Transformer and \textsc{TP-Transformer} with continuous roles on several abstractive summarization datasets, in both metrics scores and human evaluation.
We further demonstrate the syntactic structures encoded in the role vectors and show the improved syntactic interpretability in our model's hidden states.
\section{Ethics Statement}
\label{ethics}
In this work we propose a new encoder-decoder modeling architecture and build several models to benchmark our new architecture with baseline architectures on several open source summarization datasets. 

\paragraph{Intended use.}
Our architecture is designed to build models of abstractive summarization. Potentially our architecture could be used to train models for summarizing any type of company internal datasets (e.g., internal documents, reports, meetings, legal forms, etc.) to further improve the productivity and efficiency of the users in their daily activities without needing to read long documents.

\paragraph{Failure mode.}
Even though our models yield factually consistent summaries, as judged by human evaluation, they can still generate factually inconsistent summaries or sometimes hallucinate information that the source document does not include. This might be due to the bias or noise in the training data. Model builders wanting to use our architecture to build models on their company internal datasets should build models with consideration of intellectual properties and privacy rights.

\paragraph{Misuse Potential.} We note the models to be built with our architecture should be used with careful consideration. The generated summaries produced by our models are not controlled and use generative approaches, therefore, they could generate unreliable text. Researchers working on abstractive summarization should focus on generating factually correct, ethical and reliable text. If our models are trained on news datasets, a careful consideration should be made on factuality of the generated text and measures have been taken to prevent model hallucinations.  

\section*{Acknowledgments}
\vspace{-5pt}
We thank the reviewers for their helpful comments.
This work was partially supported by NSF-CAREER Award 1846185 and a Microsoft Investigator Fellowship.

\bibliography{anthology,custom}

\begin{thebibliography}{41}
\expandafter\ifx\csname natexlab\endcsname\relax\def\natexlab#1{#1}\fi

\bibitem[{Ba et~al.(2016)Ba, Kiros, and Hinton}]{ba2016layer}
Jimmy~Lei Ba, Jamie~Ryan Kiros, and Geoffrey~E Hinton. 2016.
\newblock Layer normalization.
\newblock \emph{arXiv preprint arXiv:1607.06450}.

\bibitem[{Celikyilmaz et~al.(2018)Celikyilmaz, Bosselut, He, and
  Choi}]{celikyilmaz2018}
Asli Celikyilmaz, Antoine Bosselut, Xiaodong He, and Yejin Choi. 2018.
\newblock Deep communicating agents for abstractive summarization.
\newblock In \emph{16th Annual Conference of the North American Chapter of the
  Association for Computational Linguistics: Human Language Technologies, New
  Orleans, USA.}

\bibitem[{Chen et~al.(2020)Chen, Huang, Palangi, Smolensky, Forbus, and
  Gao}]{chen2019natural}
Kezhen Chen, Qiuyuan Huang, Hamid Palangi, Paul Smolensky, Kenneth~D Forbus,
  and Jianfeng Gao. 2020.
\newblock Mapping natural-language problems to formal-language solutions using
  structured neural representations.
\newblock In \emph{Proceedings of the ICML}.

\bibitem[{Cho et~al.(2014)Cho, van Merri{\"e}nboer, Gulcehre, Bahdanau,
  Bougares, Schwenk, and Bengio}]{cho-etal-2014-learning}
Kyunghyun Cho, Bart van Merri{\"e}nboer, Caglar Gulcehre, Dzmitry Bahdanau,
  Fethi Bougares, Holger Schwenk, and Yoshua Bengio. 2014.
\newblock \href {https://doi.org/10.3115/v1/D14-1179} {Learning phrase
  representations using {RNN} encoder{--}decoder for statistical machine
  translation}.
\newblock In \emph{Proceedings of the 2014 Conference on Empirical Methods in
  Natural Language Processing ({EMNLP})}, pages 1724--1734, Doha, Qatar.
  Association for Computational Linguistics.

\bibitem[{Cohan et~al.(2018)Cohan, Dernoncourt, Kim, Bui, Kim, Chang, and
  Goharian}]{cohan-etal-2018-discourse}
Arman Cohan, Franck Dernoncourt, Doo~Soon Kim, Trung Bui, Seokhwan Kim, Walter
  Chang, and Nazli Goharian. 2018.
\newblock \href {https://doi.org/10.18653/v1/N18-2097} {A discourse-aware
  attention model for abstractive summarization of long documents}.
\newblock In \emph{Proceedings of the 2018 Conference of the North {A}merican
  Chapter of the Association for Computational Linguistics: Human Language
  Technologies, Volume 2 (Short Papers)}, pages 615--621, New Orleans,
  Louisiana. Association for Computational Linguistics.

\bibitem[{Denkowski and Lavie(2014)}]{Denkowski-14-meteor}
Michael Denkowski and Alon Lavie. 2014.
\newblock Meteor universal: Language specific translation evaluation for any
  target language.
\newblock In \emph{Proceedings of the ninth workshop on statistical machine
  translation}, pages 376--380.

\bibitem[{Devlin et~al.(2019)Devlin, Chang, Lee, and
  Toutanova}]{devlin-etal-2019-bert}
Jacob Devlin, Ming-Wei Chang, Kenton Lee, and Kristina Toutanova. 2019.
\newblock \href {https://doi.org/10.18653/v1/N19-1423} {{BERT}: Pre-training of
  deep bidirectional transformers for language understanding}.
\newblock In \emph{Proceedings of the 2019 Conference of the North {A}merican
  Chapter of the Association for Computational Linguistics: Human Language
  Technologies, Volume 1 (Long and Short Papers)}, pages 4171--4186,
  Minneapolis, Minnesota. Association for Computational Linguistics.

\bibitem[{Durmus et~al.(2020)Durmus, He, and Diab}]{durmus-etal-2020-feqa}
Esin Durmus, He~He, and Mona Diab. 2020.
\newblock \href {https://doi.org/10.18653/v1/2020.acl-main.454} {{FEQA}: A
  question answering evaluation framework for faithfulness assessment in
  abstractive summarization}.
\newblock In \emph{Proceedings of the 58th Annual Meeting of the Association
  for Computational Linguistics}, pages 5055--5070, Online. Association for
  Computational Linguistics.

\bibitem[{Gardner et~al.(2018)Gardner, Grus, Neumann, Tafjord, Dasigi, Liu,
  Peters, Schmitz, and Zettlemoyer}]{Gardner2017allennlp}
Matt Gardner, Joel Grus, Mark Neumann, Oyvind Tafjord, Pradeep Dasigi,
  Nelson~F. Liu, Matthew Peters, Michael Schmitz, and Luke Zettlemoyer. 2018.
\newblock \href {https://doi.org/10.18653/v1/W18-2501} {{A}llen{NLP}: A deep
  semantic natural language processing platform}.
\newblock In \emph{Proceedings of Workshop for {NLP} Open Source Software
  ({NLP}-{OSS})}, pages 1--6, Melbourne, Australia. Association for
  Computational Linguistics.

\bibitem[{Grusky et~al.(2018)Grusky, Naaman, and Artzi}]{grusky2018newsroom}
Max Grusky, Mor Naaman, and Yoav Artzi. 2018.
\newblock \href {http://aclweb.org/anthology/N18-1065} {Newsroom: A dataset of
  1.3 million summaries with diverse extractive strategies}.
\newblock In \emph{Proceedings of the 2018 Conference of the North American
  Chapter of the Association for Computational Linguistics: Human Language
  Technologies}, pages 708--719, New Orleans, Louisiana. Association for
  Computational Linguistics.

\bibitem[{Hermann et~al.(2015)Hermann, Kocisky, Grefenstette, Espeholt, Kay,
  Suleyman, and Blunsom}]{hermann2015teaching}
Karl~Moritz Hermann, Tomas Kocisky, Edward Grefenstette, Lasse Espeholt, Will
  Kay, Mustafa Suleyman, and Phil Blunsom. 2015.
\newblock Teaching machines to read and comprehend.
\newblock In \emph{Advances in neural information processing systems}, pages
  1693--1701.

\bibitem[{Hewitt and Manning(2019)}]{hewitt-manning-2019-structural}
John Hewitt and Christopher~D. Manning. 2019.
\newblock \href {https://doi.org/10.18653/v1/N19-1419} {{A} structural probe
  for finding syntax in word representations}.
\newblock In \emph{Proceedings of the 2019 Conference of the North {A}merican
  Chapter of the Association for Computational Linguistics: Human Language
  Technologies, Volume 1 (Long and Short Papers)}, pages 4129--4138,
  Minneapolis, Minnesota. Association for Computational Linguistics.

\bibitem[{Huang et~al.(2019)Huang, Deng, Wu, Liu, and He}]{Huang2019ATPL}
Qiuyuan Huang, Li~Deng, Dapeng Wu, Chang Liu, and Xiaodong He. 2019.
\newblock \href {https://doi.org/10.1609/aaai.v33i01.33011344} {Attentive
  tensor product learning}.
\newblock \emph{Proceedings of the AAAI Conference on Artificial Intelligence},
  33(01):1344--1351.

\bibitem[{Huang et~al.(2018)Huang, Smolensky, He, Deng, and
  Wu}]{huang-etal-2018-TPGN}
Qiuyuan Huang, Paul Smolensky, Xiaodong He, Li~Deng, and Dapeng Wu. 2018.
\newblock \href {https://doi.org/10.18653/v1/N18-1114} {Tensor product
  generation networks for deep {NLP} modeling}.
\newblock In \emph{Proceedings of the 2018 Conference of the North {A}merican
  Chapter of the Association for Computational Linguistics: Human Language
  Technologies, Volume 1 (Long Papers)}, pages 1263--1273, New Orleans,
  Louisiana. Association for Computational Linguistics.

\bibitem[{Koupaee and Wang(2018)}]{koupaee2018wikihow}
Mahnaz Koupaee and William~Yang Wang. 2018.
\newblock Wikihow: A large scale text summarization dataset.
\newblock \emph{arXiv preprint arXiv:1810.09305}.

\bibitem[{Lebanoff et~al.(2019)Lebanoff, Song, Dernoncourt, Kim, Kim, Chang,
  and Liu}]{lebanoff-etal-2019-scoring}
Logan Lebanoff, Kaiqiang Song, Franck Dernoncourt, Doo~Soon Kim, Seokhwan Kim,
  Walter Chang, and Fei Liu. 2019.
\newblock \href {https://doi.org/10.18653/v1/P19-1209} {Scoring sentence
  singletons and pairs for abstractive summarization}.
\newblock In \emph{Proceedings of the 57th Annual Meeting of the Association
  for Computational Linguistics}, pages 2175--2189, Florence, Italy.
  Association for Computational Linguistics.

\bibitem[{Liao et~al.(2018)Liao, Lebanoff, and Liu}]{liao-etal-2018-abstract}
Kexin Liao, Logan Lebanoff, and Fei Liu. 2018.
\newblock \href {https://www.aclweb.org/anthology/C18-1101} {{A}bstract
  {M}eaning {R}epresentation for multi-document summarization}.
\newblock In \emph{Proceedings of the 27th International Conference on
  Computational Linguistics}, pages 1178--1190, Santa Fe, New Mexico, USA.
  Association for Computational Linguistics.

\bibitem[{Lin(2004)}]{Lin-04-rouge}
Chin-Yew Lin. 2004.
\newblock Rouge: A package for automatic evaluation of summaries.
\newblock In \emph{Text summarization branches out: Proceedings of the ACL-04
  workshop}, volume~8. Barcelona, Spain.

\bibitem[{Lin et~al.(2019)Lin, Tan, and Frank}]{lin-etal-2019-open}
Yongjie Lin, Yi~Chern Tan, and Robert Frank. 2019.
\newblock \href {https://doi.org/10.18653/v1/W19-4825} {Open sesame: Getting
  inside {BERT}{'}s linguistic knowledge}.
\newblock In \emph{Proceedings of the 2019 ACL Workshop BlackboxNLP: Analyzing
  and Interpreting Neural Networks for NLP}, pages 241--253, Florence, Italy.
  Association for Computational Linguistics.

\bibitem[{Liu and Lapata(2018)}]{liu2018}
Yang Liu and Mirella Lapata. 2018.
\newblock Learning structured text representations.
\newblock In \emph{Transactions of the Association for Computational
  Linguistics}.

\bibitem[{Liu and Lapata(2019)}]{liu2019}
Yang Liu and Mirella Lapata. 2019.
\newblock Hierarchical transformers for multi-document summarization.
\newblock In \emph{Transactions of the Association for Computational
  Linguistics}.

\bibitem[{Manning et~al.(2014)Manning, Surdeanu, Bauer, Finkel, Bethard, and
  McClosky}]{manning-corenlp}
Christopher~D. Manning, Mihai Surdeanu, John Bauer, Jenny Finkel, Steven~J.
  Bethard, and David McClosky. 2014.
\newblock \href {http://www.aclweb.org/anthology/P/P14/P14-5010} {The
  {Stanford} {CoreNLP} natural language processing toolkit}.
\newblock In \emph{Association for Computational Linguistics (ACL) System
  Demonstrations}, pages 55--60.

\bibitem[{Maynez et~al.(2020)Maynez, Narayan, Bohnet, and
  McDonald}]{maynez-etal-2020-faithfulness}
Joshua Maynez, Shashi Narayan, Bernd Bohnet, and Ryan McDonald. 2020.
\newblock \href {https://doi.org/10.18653/v1/2020.acl-main.173} {On
  faithfulness and factuality in abstractive summarization}.
\newblock In \emph{Proceedings of the 58th Annual Meeting of the Association
  for Computational Linguistics}, pages 1906--1919, Online. Association for
  Computational Linguistics.

\bibitem[{McCoy et~al.(2019)McCoy, Linzen, Dunbar, and
  Smolensky}]{mccoy2018rnns}
R.~Thomas McCoy, Tal Linzen, Ewan Dunbar, and Paul Smolensky. 2019.
\newblock \href {https://openreview.net/forum?id=BJx0sjC5FX} {{RNN}s implicitly
  implement tensor-product representations}.
\newblock In \emph{International Conference on Learning Representations}.

\bibitem[{Nallapati et~al.(2016)Nallapati, Zhou, Gulcehre, Xiang
  et~al.}]{nallapati2016abstractive}
Ramesh Nallapati, Bowen Zhou, Caglar Gulcehre, Bing Xiang, et~al. 2016.
\newblock Abstractive text summarization using sequence-to-sequence rnns and
  beyond.
\newblock In \emph{Computational Natural Language Learning}.

\bibitem[{Narayan et~al.(2018)Narayan, Cohen, and
  Lapata}]{narayan-etal-2018-dont}
Shashi Narayan, Shay~B. Cohen, and Mirella Lapata. 2018.
\newblock \href {https://doi.org/10.18653/v1/D18-1206} {Don{'}t give me the
  details, just the summary! topic-aware convolutional neural networks for
  extreme summarization}.
\newblock In \emph{Proceedings of the 2018 Conference on Empirical Methods in
  Natural Language Processing}, pages 1797--1807, Brussels, Belgium.
  Association for Computational Linguistics.

\bibitem[{Palangi et~al.(2018)Palangi, Smolensky, He, and
  Deng}]{Palangi2018QuestionAnsweringWG}
H.~Palangi, P.~Smolensky, X.~He, and L.~Deng. 2018.
\newblock Question-answering with grammatically-interpretable representations.
\newblock In \emph{AAAI}.

\bibitem[{Pasunuru and Bansal(2018)}]{pasunuru2018}
Ramakanth Pasunuru and Mohit Bansal. 2018.
\newblock Multireward reinforced summarization with saliency and entailment.
\newblock In \emph{16th Annual Conference of the North American Chapter of the
  Association for Computational Linguistics: Human Language Technologies, New
  Orleans, USA.}

\bibitem[{Paulus et~al.(2018)Paulus, Xiong, and Socher}]{paulus2018}
Romain Paulus, Caiming Xiong, and Richard Socher. 2018.
\newblock A deep reinforced model for abstractive summarization.
\newblock In \emph{6th International Conference on Learning Representations,
  Vancouver, BC, Canada.}

\bibitem[{Raffel et~al.(2020)Raffel, Shazeer, Roberts, Lee, Narang, Matena,
  Zhou, Li, and Liu}]{raffel2020T5}
Colin Raffel, Noam Shazeer, Adam Roberts, Katherine Lee, Sharan Narang, Michael
  Matena, Yanqi Zhou, Wei Li, and Peter~J Liu. 2020.
\newblock Exploring the limits of transfer learning with a unified text-to-text
  transformer.
\newblock \emph{Journal of Machine Learning Research}, 21(140):1--67.

\bibitem[{Schlag and Schmidhuber(2018)}]{schlag2018tprrnn}
Imanol Schlag and J{\"u}rgen Schmidhuber. 2018.
\newblock Learning to reason with third order tensor products.
\newblock In \emph{Advances in Neural Information Processing Systems}, pages
  10002--10013.

\bibitem[{Schlag et~al.(2019)Schlag, Smolensky, Fernandez, Jojic, Schmidhuber,
  and Gao}]{schlag2019enhancing}
Imanol Schlag, Paul Smolensky, Roland Fernandez, Nebojsa Jojic, J{\"u}rgen
  Schmidhuber, and Jianfeng Gao. 2019.
\newblock Enhancing the transformer with explicit relational encoding for math
  problem solving.
\newblock \emph{arXiv preprint arXiv:1910.06611}.

\bibitem[{See et~al.(2017)See, Liu, and Manning}]{see-etal-2017-get}
Abigail See, Peter~J. Liu, and Christopher~D. Manning. 2017.
\newblock \href {https://doi.org/10.18653/v1/P17-1099} {Get to the point:
  Summarization with pointer-generator networks}.
\newblock In \emph{Proceedings of the 55th Annual Meeting of the Association
  for Computational Linguistics (Volume 1: Long Papers)}, pages 1073--1083,
  Vancouver, Canada. Association for Computational Linguistics.

\bibitem[{Shazeer and Stern(2018)}]{adafactor-shazeer18a}
Noam Shazeer and Mitchell Stern. 2018.
\newblock Adafactor: Adaptive learning rates with sublinear memory cost.
\newblock In \emph{International Conference on Machine Learning}, pages
  4596--4604. PMLR.

\bibitem[{Smolensky(1990)}]{smolensky1990tpr}
Paul Smolensky. 1990.
\newblock Tensor product variable binding and the representation of symbolic
  structures in connectionist systems.
\newblock \emph{Artificial intelligence}, 46(1-2):159--216.

\bibitem[{Song et~al.(2018)Song, Zhao, and Liu}]{song-etal-2018-structure}
Kaiqiang Song, Lin Zhao, and Fei Liu. 2018.
\newblock \href {https://www.aclweb.org/anthology/C18-1146} {Structure-infused
  copy mechanisms for abstractive summarization}.
\newblock In \emph{Proceedings of the 27th International Conference on
  Computational Linguistics}, pages 1717--1729, Santa Fe, New Mexico, USA.
  Association for Computational Linguistics.

\bibitem[{Soulos et~al.(2019)Soulos, McCoy, Linzen, and
  Smolensky}]{soulos2019discovering}
Paul Soulos, Tom McCoy, Tal Linzen, and Paul Smolensky. 2019.
\newblock Discovering the compositional structure of vector representations
  with role learning networks.
\newblock \emph{arXiv preprint arXiv:1910.09113}.

\bibitem[{Tenney et~al.(2019{\natexlab{a}})Tenney, Das, and
  Pavlick}]{tenney-etal-2019-bert}
Ian Tenney, Dipanjan Das, and Ellie Pavlick. 2019{\natexlab{a}}.
\newblock \href {https://doi.org/10.18653/v1/P19-1452} {{BERT} rediscovers the
  classical {NLP} pipeline}.
\newblock In \emph{Proceedings of the 57th Annual Meeting of the Association
  for Computational Linguistics}, pages 4593--4601, Florence, Italy.
  Association for Computational Linguistics.

\bibitem[{Tenney et~al.(2019{\natexlab{b}})Tenney, Xia, Chen, Wang, Poliak,
  McCoy, Kim, Durme, Bowman, Das, and Pavlick}]{tenney2018edge}
Ian Tenney, Patrick Xia, Berlin Chen, Alex Wang, Adam Poliak, R~Thomas McCoy,
  Najoung Kim, Benjamin~Van Durme, Sam Bowman, Dipanjan Das, and Ellie Pavlick.
  2019{\natexlab{b}}.
\newblock \href {https://openreview.net/forum?id=SJzSgnRcKX} {What do you learn
  from context? probing for sentence structure in contextualized word
  representations}.
\newblock In \emph{International Conference on Learning Representations}.

\bibitem[{Vaswani et~al.(2017)Vaswani, Shazeer, Parmar, Uszkoreit, Jones,
  Gomez, Kaiser, and Polosukhin}]{vaswani2017attention}
Ashish Vaswani, Noam Shazeer, Niki Parmar, Jakob Uszkoreit, Llion Jones,
  Aidan~N Gomez, {\L}ukasz Kaiser, and Illia Polosukhin. 2017.
\newblock Attention is all you need.
\newblock In \emph{Advances in neural information processing systems}, pages
  5998--6008.

\bibitem[{Zhang et~al.(2020)Zhang, Zhao, Saleh, and Liu}]{zhang2020pegasus}
Jingqing Zhang, Yao Zhao, Mohammad Saleh, and Peter~J Liu. 2020.
\newblock Pegasus: Pre-training with extracted gap-sentences for abstractive
  summarization.
\newblock In \emph{Proceedings of the 37 th International Conference on Machine
  Learning}.

\end{thebibliography}
\bibliographystyle{acl_natbib}

\appendix

\section*{Appendix}

\section{\textsc{TP-Transformer} Architecture}
\begin{figure}[ht!]
\begin{center}
\includegraphics[width=0.4\textwidth]{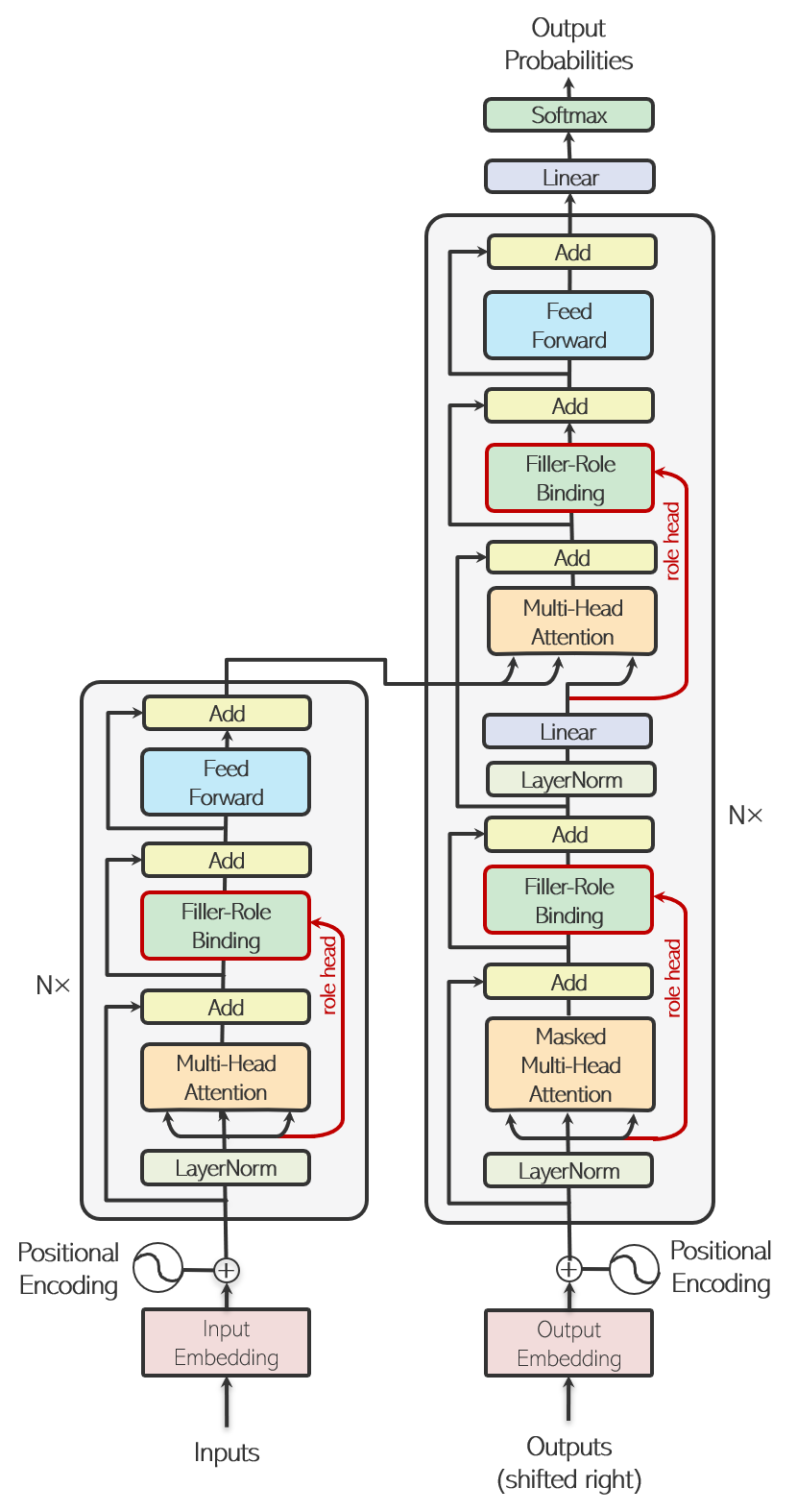}
\end{center} 
\caption{\small \textsc{TP-Transformer} model architecture.}
\label{fig:entire_model}
\end{figure}
We provide \figref{fig:entire_model} to visualize the full encoder-decoder architecture of our \textsc{TP-Transformer}.

\section{Summarization Experimental Setup}
The Transformer and the two \textsc{\textsc{TP-Transformers}} all have 6 layers, 8 heads per layer, dimension per head $d_k$=64, model dimension $d_m$=512, and feed-forward dimension $d_f$=2048 for the encoder and decoder.
Our \textsc{TP-Transformer} with discrete roles has $N_r$=50 role embeddings of dimension $d_r=64$ at every layer.
We search the optimal $N_r$ from $\{20, 50, 100, 200, 1000\}$ and select the one with the best validation set performance. 
For each of the dataset above, we train the all three models from scratch using an Adafactor Optimizer~\cite{adafactor-shazeer18a} with square root learning rate decay and dropout rate of 0.1.
The total number of parameter of the Transformer, \textsc{TP-Transformer} with continuous roles, and our \textsc{TP-Transformer} with discrete roles are 60506880, 65234688, and 64258080 respectively. 
Every model is trained on 4 NVidia V100 GPUs (32GB) with a batch size of 32 per GPU.

\subsection{Human Evaluation}
We conduct human evaluation to compare the summaries generated by the original Transformer and our \textsc{TP-Transformer}.
We randomly sample 120 examples from the test sets of XSum and Wikihow datasets with the corresponding beam-searched model summaries.
For every example, we show the source document, the reference summary, and two model summaries shuffled in order to three human evaluators, and ask them to decide which summary is better in six different aspects: grammar, coherency, factuality, saliency, redundancy, and an overall preference.
We then take the majority vote of every examples from its three human annotators.

\section{Probing Experimental Setup}
As there is no existing dataset for probing decoders, we create our own training and evaluation data by running off-the-shelf models on the summarization datasets.
Specifically, to probe a decoder trained on the XSum dataset on the POS task, we run an POS tagger on the reference summaries from the XSum training set and the model-generated summaries for the XSum dev set to create the ground-truth labels for the training set and model-specific dev set. 
We use Stanford CoreNLP~\cite{manning-corenlp} to get the labels for POS, dependency and named entity probing tasks. 
We use a BERT-base model~\cite{devlin-etal-2019-bert} from AllenNLP~\cite{Gardner2017allennlp} to get the ground-truth labels for SRL.
We restore the model trained on a summarization dataset and freeze its parameters during the probing. 
We simply add a linear layer on top of the target representation to project it onto the output label space.

\section{Related Works}
\paragraph{Implicit TPR Encodings in Neural Networks}
\citet{mccoy2018rnns} showed that, in GRU-based~\cite{cho-etal-2014-learning} encoder-decoder networks performing fully-compositional string manipulations, trained on extensive data that fully exemplifies the range of possible compositions, the medial encoding between encoder and decoder could be extremely well approximated by TPRs.
\citet{soulos2019discovering} presented the ROLE model that learns its own role scheme to optimize the fit of a TPR approximation to a given set of internal representations in a pre-trained target neural network, removing the need for human-generated hypotheses about the role schemes the network might be implementing.
While this work successfully interprets the Tensor Product Representation in fully compositional tasks, abstractive summarization, as well as most other NLP tasks, are only partially compositional and the symbolic rules in language are much more complex.
Although these two works showed that Tensor Product Representation can naturally emerge in a unstructured representations, we argue that standard models only learn TPRs without any special bias to do so when the compositional structure of the task is simple and blatant and when the training set makes that painfully clear by providing a good sample of the compositional possibilities. 
That is possible for the simple string tasks addressed in the two previous works, but not in the abstractive summarization as well as other real-world NLP tasks, where we show that having explicit TPR helps in modeling the structure information. 

\paragraph{Sequence Models Encode Implicit Structure.}
Several recent works have shown that the pretrained Transformer-based BERT~\cite{devlin-etal-2019-bert} embeddings implicitly encode structural linguistic relations with various interpretation methods.
The first, and also the most popular method~\cite{tenney-etal-2019-bert} is to train an auxiliary classifier to probe the model's hidden representations for specific linguistic information.
The second method~\cite{lin-etal-2019-open} abstracts the Transformer model into a graph based on the attention weights, and explores syntactic structures based on the graph's structure.
The third method~\cite{hewitt-manning-2019-structural} sees the hidden representations of BERT as in a metric space and directly connect the distance between representations to the distance between elements in a symbolic structure (e.g., a dependency-parse tree) to extract the implicit structures without extra training. 
The interpretation method deployed here falls under the probing family, but future work will also pursue other interpretation methods.

\section{Examples of Generated Summary}
We provide examples generated by the Transformer baseline and our \textsc{TP-Transformer} in \tabref{table:example_summ_1} and \tabref{table:example_summ_2}.

\begin{table*}[t]
\centering
\begin{scriptsize}
\begin{tabular}{p{1.6cm}|p{13.2cm}}
\toprule
 \centering Datasets & Summary \\
\midrule
\centering \multirow{20}{*}{\centering Source} & Nottinghamshire Police said it would expand its categories to include misogynistic incidents.It means abuse or harassment which might not be a crime can be reported to and investigated by the police, and support for the victim put in place.Nottingham Women's Centre said it hopes it will help give more victims the courage to report incidents.Chief Constable Sue Fish claimed it will make the county a safer place for women. </br>"What women face, often on a daily basis, is absolutely unacceptable and can be extremely distressing," she said. </br>"Nottinghamshire Police is committed to taking misogynistic hate crime seriously and encourages anyone who is affected by it to contact us without hesitation. </br>"Work on the idea first started with the Nottinghamshire Safer for Women Conference last year, co-hosted by the police with the Nottingham Women's Centre.BBC TV reporter Sarah Teale was harassed in the street while reporting on the conference.The force defines misogyny hate crime as: "Incidents against women that are motivated by an attitude of a man towards a woman and includes behaviour targeted towards a woman by men simply because they are a woman. </br>"The classification now means people can report incidents which might not be considered to be a crime and the police will investigate.Nottingham Women's Centre has been helping train call centre, force control staff and officers on the beat to recognise misogynistic hate crime and ways to tackle it.These officers will also examine if and how a victim can be supported or if anything can be done to help prevent them being targeted again.Domestic abuse will not be recorded as a misogyny hate crime because it has its own procedure, the force said.Melanie Jeffs, centre manager at Nottingham Women's Centre, said: "We're pleased to see Nottinghamshire Police recognise the breadth of violence and intimidation that women experience on a daily basis in our communities. </br>"She added: "Recording this as a hate crime will give us a detailed picture of how often, when and where it is happening. </br>It has been very difficult to build that picture before but we will now get detailed data to analyse. </br>"Showing that the police take it seriously will also give people the confidence to come forward and report offences. </br>"A crime that the victim or any other person perceives to be motivated by hostility or prejudice towards any aspect of a person's identity.Police forces in England, Wales and Northern Ireland annually monitor five strands of hate crime:Forces can include their own definition of a hate crime with several recently adding sub cultures. \\
\midrule
 \centering Reference & Harassment of women is to be recorded as a hate crime in a bid to tackle sexist abuse. \\
\midrule
\centering Transformer & Women who commit misogyny and harassed a woman are to be asked to take part in an anti-Semitic conference. \\
\midrule
 \centering \textsc{TP-Transformer} & A police force has launched a national drive to combat misogyny and hate crimes in Nottinghamshire. \\
\bottomrule
\end{tabular}
\caption{\small An example from the XSum dev set and the summaries generated by the Transformer baseline and \textsc{TP-Transformer}.
}
\label{table:example_summ_1}
\end{scriptsize}
\end{table*}

\begin{table*}[t]
\centering
\begin{scriptsize}
\begin{tabular}{p{1.6cm}|p{13.2cm}}
\toprule
 \centering Datasets & Summary \\
\midrule
\centering \multirow{14}{*}{\centering Source} & Sixty patrol boats will protect the UK's two new aircraft carriers which are due to arrive at Portsmouth Naval Base in 2017.The first carrier, HMS Queen Elizabeth, is expected to be operational in 2020. </br>"We are going to see a bigger Royal Navy and the flagship... will be here in Portsmouth," Michael Fallon said.The 60 Pacific 24 rigid-hulled inflatable boats will be built by BAE systems to "guard the carriers in the harbour and our new frigates and destroyers", Mr Fallon said.He said they will also enhance security by providing a rapid response in rescue, anti-piracy and counter-narcotics missions in the area.Mr Fallon said: "Through the defence review, defence spending is going to go up every April for the rest of this parliament.He said as part of the larger investment, the government will also be able to provide the new aircraft carriers with sufficient fighter jets. </br>"We have said we will maintain a minimum fleet of 19 destroyers and frigates, but as the older frigates are retired we also hope to add a lighter frigate between the offshore patrol vessel and Type 26 and to build more of those as well. </br>"Mr Fallon's visit to Portsmouth Naval Base comes as work has begun to rebuild the jetty for the arrival of HMS Queen Elizabeth in 2017.Floating cranes are also dredging Portsmouth harbour to prepare deeper channels for the aircraft carriers to sail from the base, which are the largest ships ever built for the Royal Navy. </br>"This is a huge financial investment in making sure the channel is wide enough, in enlarging the jetty here so they can take the carriers and in making sure the carriers are properly guarded," Mr Fallon said.Taller than Nelson's Column and longer than Portsmouth's Spinnaker Tower laid on its side, the new carriers will displace 65,000 tonnes of water.To make room for the carriers three million cubic metres of clay, sand and gravel will be removed from a two-mile stretch of Portsmouth Harbour covering an area the size of 200 football pitches. \\
\midrule
 \centering \multirow{2}{*}{\centering Reference} & Increased spending will result in a "bigger" Royal Navy, the defence secretary has said, as he announced a new £13.5m shipbuilding contract. \\
\midrule
\centering Transformer & The Royal Navy's new aircraft carriers will be patrolling the Portsmouth harbour this year, the defence secretary has said. \\
\midrule
 \centering \textsc{TP-Transformer} & Plans for a new Royal Navy aircraft carriers to be built in Portsmouth have been unveiled. \\
\bottomrule
\end{tabular}
\caption{\small An example from the XSum dev set and the summaries generated by the Transformer baseline and \textsc{TP-Transformer}.
}
\label{table:example_summ_2}
\end{scriptsize}
\end{table*}

\end{document}